\relax
\documentclass[letterpaper]{article} 
\usepackage{aaai19}  
\usepackage{times}  
\usepackage{helvet}  
\usepackage{courier}  
\usepackage{url}  
\usepackage{graphicx}  
\usepackage{booktabs} 
\usepackage{amsmath}
\usepackage{amssymb}
\usepackage{latexsym}
\usepackage{algpseudocode}
\usepackage{algorithm}
\usepackage{algorithmicx}
\usepackage{bm}
\usepackage{subfigure}

\newtheorem{assump}{\textbf{Assumption}}
\newtheorem{theorem}{\textbf{Theorem}}

\newtheorem{lemm}{\textbf{Lemma}}
\newtheorem{remark}{\textbf{Remark}}

\newcommand{\argmin}{arg\,min}\newcommand{\nn}{\nonumber}

\begin{document}
%
\title{Active Learning in Recommendation Systems with Multi-level User Preferences}
\author{Yuheng Bu \thanks{This work was done during an internship at Amazon. University of Illinois at Urbana-Champaign. Email: bu3@illinois.edu}
\And  Kevin Small  \thanks{Amazon Inc., Email: smakevin@amazon.com}}

\maketitle
\begin{abstract}
While recommendation systems generally observe user behavior passively, there has been an increased interest in directly querying users to learn their specific preferences. In such settings, considering queries at different levels of granularity to optimize user information acquisition is crucial to efficiently providing a good user experience. In this work, we study the active learning problem with multi-level user preferences within the collective matrix factorization (CMF) framework. CMF jointly captures multi-level user preferences with respect to items and relations between items (e.g., book genre, cuisine type), generally resulting in improved predictions. Motivated by finite-sample analysis of the CMF model, we propose a theoretically optimal active learning strategy based on the Fisher information matrix and use this to derive a realizable approximation algorithm for practical recommendations. Experiments are conducted using both the Yelp dataset directly and an illustrative synthetic dataset in the three settings of personalized active learning, cold-start recommendations, and noisy data --  demonstrating strong improvements over several widely used active learning methods.
\end{abstract}

\section{Introduction}
\label{sec:intro}

Recommendation systems are widely studied in both academic and commercial settings. Most existing work considers the {\em passive, item-level feedback} scenario where the recommendation system observes historical user feedback for a set of items with respect to a population of users and estimates unobserved user-item utility to make recommendations. However, exclusively considering item-level feedback disregards frequently available information regarding relations (i.e., cuisine type, product substitutability~\cite{McAuleyPL15}) between entities (i.e., items, users, cuisines, etc.) in the database. Additionally, standard recommendation systems only observe historical user-item responses to estimate model parameters, unlike {\em active learning} systems that are able to directly query the user to more efficiently learn user preferences. Noting these shortcomings, online recommendation systems~\cite{BreslerCS14}, preference elicitation methods~\cite{ChenK06}, and interactive recommendation systems~\cite{MahmoodR09} all provide directions to mitigate these issues.

In this work, we are specifically interested in adding system-initiative capabilities with respect to querying the user. As motivation, suppose you are interested in purchasing book to read during an upcoming vacation.  A standard recommendation system would observe your past purchases/ratings and select a book expected to align with learned preferences.  However, a multi-turn dialogue agent may be able to ask preference questions of varying types and levels of granularity (i.e., {\em Did you like G.K. Chesterton's `The Man Who Was Thursday'?} [item utility elicitation], {\em Are you looking for light reading?} [use case information], {\em Do you like science fiction novels?} [category information]).  These query families result in greater flexibility than simply making recommendations and allow asking questions that can more quickly improve the model for making session-specific recommendations.

The core of our proposed method is an active learning extension to the collective matrix factorization (CMF) model. CMF produces a low-dimensional embedding that is shared across each relation for which the item/user participates and jointly represents all available sources of information on different levels~\cite{gupta2015collective}.  By enabling active learning within CMF, we can generate a personalized active learning session to efficiently estimate the CMF parameters and make high-quality recommendations.
%
Our primary contributions include: (1) framing the question selection problem for multi-level user preferences in a system-initiative active recommendation system within the CMF framework (2) providing a theoretical analysis of an optimal active learning strategy for CMF and corresponding realizable approximation and (3) demonstrating that the proposed algorithm outperforms strong baselines on real-world Yelp data and an illustrative synthetic dataset that explicitly satisfies the CMF assumptions in standard, cold start, and noisy data settings.








\section{Related Works}
\label{sec:related_work}
This work draws upon several existing research areas. Below, we itemize some of the most relevant related work.

\noindent
{\bf Recommendation Systems}:
Recommendation systems are frequently categorized as content-based~\cite{PazzaniB07} or collaborative filtering~\cite{KorenBV09} methods. This work builds upon the collective matrix factorization (CMF) model~\cite{singh2008relational,gupta2015collective}, a collaborative filtering method that generalizes matrix factorization to also account for different levels of relations between entities. Within the collaborative filtering approach, observation sparsity is the primary limitation, particularly in the cold start setting (i.e., where a new item has a small number of ratings or a new user has rated a small number of items). Approaches to ameliorate this issues include user preference elicitation~\cite{RashidKR08}, interview construction~\cite{ZhouYZ11,SunLLKLZ13}, and optimal experimental design~\cite{anava2015budget}.  The distinguishing aspect of our work is that we develop an active and personalized querying strategy that jointly accounts for several types of questions based on item preferences and other relationships between entities.

\noindent
{\bf Active Learning}:
Active learning has been widely studied (e.g., \cite{Settles10} for a general survey, \cite{RubensKS11,elahi2016survey} in the context of recommendation systems), describing when a learning algorithm observes a large set (or stream) of unlabeled examples and can choose a subset for labeling -- attempting to maximize performance while minimizing annotation effort. The most closely related work within the recommendation systems setting is active learning in the matrix completion scenario~\cite{BhargavaGN17}. While related, this differs from our work as we use the CMF setting to jointly estimate a representation for several levels of relationships.  From a methodological perspective, our work draws upon recent results for active learning with Fisher information based querying functions~\cite{SouratiALED17} and convergence properties of active learning for maximum likelihood estimation~\cite{ChaudhuriKNS15}.  We expand upon these results for the CMF setting.

\noindent
{\bf Conversational Agents}:
One potential motivation for this work is conversational recommendations~\cite{MahmoodR09,christakopoulou2016towards}, where questions are determined by modeling different relationship types within a unified framework and a theoretically well-motivated active learning strategy. However, it should be noted that developing goal-oriented conversational agents (e.g., \cite{Young13}) and specifically dialogue managers is a very mature AI subfield. We are only considering the restricted setting of asking a personalized set of questions to provide an optimized recommendation.

\section{Preliminaries and Model}
\label{sec:model}
We first briefly review the probabilistic collective matrix factorization model~\cite{singh2008relational} 
and concretely formalize our active learning extension.

\subsection{General Notation}
We use lower case letters to denote scalars and vectors, upper case letters to denote matrices and sets. $I$ is an appropriately sized identity matrix. Superscript $^T$ denotes a vector or matrix transpose and $|\cdot|$ denotes the support size of a set. The $\ell_2$-norm of a vector $x \in \mathbb{R}^k$ is defined as $\|x \|_2=\sqrt{\sum_{i=1}^k x_i^2}$ where $\|x\|_A=\sqrt{x^T A x}$ is defined for a vector $x$ and a matrix $A$ of appropriate dimensions. The Frobenius norm of a matrix $A \in \mathbb{R}^{m\times n}$ is defined as $\|A\|_F=\sqrt{\sum_{i=1}^m \sum_{j=1}^n |A_{ij}|^2}$.

\subsection{Relational Data}
We represent the set of entities by $\mathcal{E}$ and the set of relations between them by $\mathcal{R}$, respectively. Denote the observed database by $\mathcal{D}$, which consists of tuples of the form $\{r_i, e^{(1)}_{i}, e^{(2)}_{i}, y_i\}^N_{i=1}$, where $r_i \in \mathcal{R}$ is a relation between two different types of entities, $ e^{(1)}_{i}, e^{(2)}_{i} \in \mathcal{E}$ are a pair of different entities, $y_i \in \{\pm 1\}$ is the label denoting whether $r_i(e^{(1)}_{i},e^{(2)}_{i})$ holds (or not), and $N$ is the total number of observed relations in database $\mathcal{D}$.

For example, a simple database that consists only of the user ratings for restaurants would contain users and businesses as the entities, and only a single relation $r\in \mathcal{R}$, such that $r(e^{(U)}_{i},e^{(B)}_{i}) = 1$ if the user $e^{(U)}_{i}$ liked the business $e^{(B)}_{i}$.
As in this example, many real-life databases are sparse (i.e., only a very small subset of possible relations are observed) and the goal of modeling is to be able to complete this database such that we can make recommendation based on the prediction. Specifically, given any query $r_q(e^{(1)}_{q}, e^{(2)}_{q})$ that is absent from the observed database, we would like to predict whether the relation holds.

\subsection{Collective Factorization Model}
The collective matrix factorization model \cite{singh2008relational,gupta2015collective} extends the commonly used matrix factorization model to multiple matrices by assigning each entity a low-dimensional latent vector that is shared across all relations where the entity appears. Formally, we assign each entity $e\in \mathcal{E}$ in our database a $k$-dimensional latent vector $\phi_e \in \mathbb{R}^k$, and denote the matrix of all such latent vectors by $\Phi \in \mathbb{R}^{k \times |\mathcal{E}|}$. We model the probability that $r_i(e^{(1)}_i,e^{(2)}_i)$ equals to $y_i$ by:
\begin{equation}\label{eq:CMF}
\mathbb{P}_\Phi[r_i(e^{(1)}_i,e^{(2)}_i) = y_i] = \sigma(y_i\cdot\phi^{(1)T}_{e_i}\cdot \phi^{(2)}_{e_i}),
\end{equation}
where $\sigma$ is the sigmoid function $\big(\mathrm{i.e.,} \, \sigma(s)=\frac{1}{1+e^{-s}}\big)$. 

The CMF model presents a number of advantages in our setting. By sharing the entity factors amongst all the relations, we are able to produce a joint representation based on all
sources of information on different levels. For example, the factors used predict user ratings will leverage information from other ratings in a collaborative filtering fashion,
from business categories via the set inclusion relationships, and even from words that appear in the reviews (the details of the model as applied to the Yelp data are described in Section \ref{sec:yelp}). By developing this joint representation, it also allows the active learning algorithm to query different type of personalized questions to different users in multi-turn recommendation system settings. A further advantage of learning
CMF model is that all the entities are effectively embedded in the same $k$-dimensional space, and thus similarities
and distances can be computed and analyzed for any set of entities.
Finally, test-time inference takes constant time and thus is very efficient: we only require a dot-product between
low-dimensional vectors for estimating the probability of a relation existing between a pair of entities.

\subsection{Problem Definition}
\label{subsec:problem}
In defining our personalized active learning problem, let $u \in S^{(U)}$ be a chosen user, where $S^{(U)}\subseteq \mathcal{E}$ is the set of all user entities. Let $S^{(E)}\subseteq \mathcal{E}$ is the set of all entities except for user entities and $T_u \subseteq S^{(E)}$ represent the set of entities, where the label of relations between user $u$ and entities are observed in the database $\mathcal{D}$, and denote $S_u \subseteq S^{(E)}$ as the set of entities where the label of their relations with user $u$ are missing in the database $\mathcal{D}$. We assume the active learning procedure operates in $T$ iterations. In each iteration, the active learning algorithm will choose $M$ entities (denoted the set by $Q_u$) from the unlabeled set $S_u$, and get answers by asking user questions with respect to these $M$ entities -- adding these labeled entities into the existing labeled set $T_u$ and removing $Q_u$ from the unlabeled entities set $S_u$. The goal of active learning is to find $M \cdot T$ entities in total, specifically the most informative questions, to query their relations with the current user $u$.


More precisely, in the $t$-th iteration, we have a labeled entity set $T_u$ and unlabeled entity set $S_u$. Given a constraint on the number of questions $M$, our goal is to select $M$ questions for user $u$ in order to minimize the cross entropy loss on the set of the unlabeled entities $S_u $.

If we define $  \ell(y|\hat{\phi}_e, \hat{\phi}_u ) \triangleq -\log \sigma(y \cdot \hat{\phi}_e^T \cdot \hat{\phi}_u )$ to be the negative log likelihood function of the CMF model with parameters $\hat{\Phi}$ estimated by our algorithm. And assume that the database is generated from CMF
model with some unknown true parameter $\Phi^*$.  Then, the cross entropy loss for a specific user $u$ can be written as,
\begin{align}\label{eq:user-cross-entropy}
L_{S_u}(\hat{\Phi}) \triangleq \mathbb{E}_{\Phi^*}\Big[\frac{1}{|S_u|}\sum_{e \in S_u} \ell(Y| \hat{\phi}_e,\hat{\phi}_u) \Big],
\end{align}
where $\mathbb{E}_{\Phi^*}$ denotes taking expectation of $Y$ over the CMF distribution with $\Phi^*$. 

Our goal is to minimize the following objective function:
\begin{equation}\label{eq:objective}
  \min_{\hat{\Phi},Q_u} \mathbb{E}\Big[ L_{S_u \setminus Q_u}(\hat{\Phi})\Big],
\end{equation}
where the expectation is taken over the estimation of $\hat\Phi$.

Clearly, the objective function inherently depends on how we estimate the latent matrix $\hat{\Phi}$ given the relations from the observed set $T_u$ and the answers of active learning questions $Q_u$ provided by user. Thus we can decompose the problem in Equation \eqref{eq:objective} into two distinct problems:

\begin{enumerate}
  \item \textbf{Model estimation}: given a subset of entities $T_u$ and their relations with user $u$, estimate the matrix $\hat{\Phi}$, which minimizes the cross entropy loss in \eqref{eq:user-cross-entropy}.
  \item \textbf{Question Selection}: given a pool of available questions $S_u$, select the most informative subset of $S_u$ as the active learning questions $Q_u$.
\end{enumerate}

\section{Algorithm and Analysis}
\label{sec:active}

In this section we tie together the model estimation and question selection problems, and present an approach that tackles the
unified problem as a whole. We derive a finite-sample estimation error bound, which is the theoretical underpinning motivating the optimal active learning strategy proposed in this section.


Essentially, the optimality relies on the assumption that the labels are generated via CMF model defined in \eqref{eq:CMF}, and consequently on the method used for model estimation: maximum likelihood (ML) estimation. To make this point clearer we first introduce the ML estimator.

\subsection{Maximum Likelihood Estimation}
Recall that we assume the database is generated from CMF model with some unknown parameter $\Phi^*$, it is impossible for us to compute the loss function \eqref{eq:user-cross-entropy} directly.
However, we can compute the empirical loss for user $u$ on labeled set $T_u$ by taking average on the observed data as follows
\begin{equation}\label{eq:emprical-user-CE}
  \hat{L}_{T_u}(\hat{\Phi}) \triangleq \frac{1}{|T_u|}\sum_{e \in T_u} \ell ({y_{eu}|\hat{\phi}_e , \hat{\phi}_u}),
\end{equation}
where $y_{eu}$ denotes the observed label of the relation between user $u$ and other entity $e$.


To jointly estimate the latent vectors for all users and other entities, we define the empirical cross-entropy loss on the entire database $\mathcal{D}$ as follows,
\begin{equation}
\hat{L}_{\mathcal{D}}(\hat{\Phi}) 
 \triangleq  \frac{1}{N} \sum_{u \in S^{(U)}}\sum_{e \in T_u} \ell(y_{eu}|\hat{\phi}_e,\hat{\phi}_u).
\end{equation}




To avoid over-fitting, we place a prior on all latent vectors in the form of a zero-mean Gaussian distribution with identity covariance matrix, i.e., $\mathcal{N}(0, 1/\lambda I)$. Minimizing the following $\ell_2$ regularized empirical loss on the observed database $\mathcal{D}$ gives the maximum likelihood estimator of the latent matrix $\Phi$,
\begin{equation}\label{eq:MLE_phi}
\hat{\Phi}_{\mathrm{ML}}=\mathrm{ argmin}_{\Phi\in \Lambda^{k \times |\mathcal{E}|}} \Big( \hat{L}_{\mathcal{D}}(\Phi) + \lambda \|\Phi\|_F^2 \Big),
\end{equation}
where $\lambda$ is a positive regularization parameter and $\Lambda \subseteq \mathbb{R}^k$ is a compact subset of $\mathbb{R}^k$.

As with several matrix factorization implementations, we use stochastic gradient descent (SGD) by cycling over the entries of the database multiple times, updating the latent factors in the direction of stochastic gradient for each entry.

\subsection{Active Learning Based on Error Bound Minimization}
\label{subsec:bound}



As discussed in Appendix \ref{appx:assum}, the maximum likelihood estimator in \eqref{eq:MLE_phi} is consistent and asymptotic normal. Since there are multiple users related with the same entity (ratings from different users), we can treat the ML estimates $\hat{\phi}_{e \mathrm{ML}}$ in \eqref{eq:MLE_phi} as a good estimation for latent vectors $\phi_e,\ e\in S_u$, and use active learning techniques to improve the estimation of $\phi_u$. Our discussion below will focus on determining the best strategy to choose $\phi_e$ actively, such that the optimal convergence rate for estimating $\phi_u$ can be achieved.

Given an estimation of $\phi_e$, $L_{S_u}(\hat\Phi), \hat{L}_{T_u}(\hat\Phi)$ defined in \eqref{eq:user-cross-entropy} and \eqref{eq:emprical-user-CE} are exclusively functions of $\hat\phi_u$. Thus, we will use the notation $L_{S_u}(\hat\phi_u), \hat{L}_{T_u}(\hat\phi_u)$ in the following discussion. For our active learning to work correctly, we require the following conditions.
\begin{assump}\label{assump:fisher}
For any $\phi_e, \phi_u \in \Lambda \subseteq \mathbb{R}^k, y \in \{\pm 1\}$, the Hessian matrix of the log likelihood function
\begin{equation}
H(\phi_e,\phi_u)\triangleq\frac{\partial^2 \ell(y|\phi_e,\phi_u)}{\partial \phi_u^2}
\end{equation}
 is a function of only
$\phi_e$ and $\phi_u$ (does not depend on $y$.)
\end{assump}

Then, the Fisher information matrix of parameter $\phi_u$ on set $S_u$ can be written as: $I_{S_u}(\phi_u)\triangleq \frac{1}{|S_u|}\sum_{e \in S_u}H(\phi_e,\phi_u)$, which is not a function of the labels $y_{eu}$.
In addition to above, we need the following assumption holds to establish the optimality of the active question selection algorithm.
\begin{assump}\label{assump:opt}
(Active learning regularity conditions)
\begin{enumerate}
  \item \textbf{Concentration at $\phi_u^*$}: For any $e \in S_u$ and $y$, we have
  \begin{align}
    &\Big\|\nabla\ell(y|\phi_e,\phi_u^*)\Big\|_{I_{S_u}(\phi_u^*)^{-1}}\le L_1,\quad \mbox{and}\nn \\
    &\Big\|I_{S_u}(\phi_u^*)^{-1/2} H(\phi_e,\phi^*_u) I_{S_u}(\phi_u^*)^{-1/2}\Big\|\le L_2
  \end{align}
  holds with probability one.
  \item \textbf{Lipschitz continuity}: There exists a neighborhood $B$ of $\phi_u^*$ and a constant $L_3$, such that for all $e \in S_u$, $H(\phi_e,\phi_u)$ are $L_3$-Lipschitz, namely,
    \begin{align}
      \Big\|I_{S_u}(\phi_u^*)^{-1/2}\big(H(\phi_e,\phi_u)- & H(\phi_e,\phi_u')\big)I_{S_u}(\phi_u^*)^{-1/2}\Big\| \nn \\
                    &\le L_3\|\phi_u-\phi_u'\|_{I_{S_u}(\phi_u^*)}
    \end{align}
    holds, for $\phi_u,\phi_u' \in B$.
\end{enumerate}
\end{assump}


We now present the main result of our paper. The proof of the following theorem and all the
supporting lemmas will be presented in Appendix \ref{appx:lemmas} and \ref{appx:thm1}.


\begin{theorem}\label{thm:main}
Suppose $\ell$ satisfies the regularity conditions in Assumptions \ref{assump:fisher}, \ref{assump:opt} and \ref{assump:LAN}. Let $\hat{\phi}_{Q_u}$ be the ML estimate using question set $Q_u$
\begin{equation}\label{eq:opt_estimator}
  \hat{\phi}_{Q_u}= \argmin_{\phi_u} \hat{L}_{Q_u}( \phi_{u} )+ \lambda \|\phi_u\|_2^2.
\end{equation}
Suppose further that $I_{Q_u}(\phi_u^*) \succeq c I_{S_u}(\phi_u^*)$ for some constant $c$. Then, $M$ large enough such that $\epsilon_M \triangleq \mathcal{O}\Big(\frac{1}{c^2}(L_1L_3+\sqrt{L_2})\sqrt{\frac{ \log k M}{M}})\Big)<1$, we have:
\begin{align}
  (1-\epsilon_M)\frac{\tau^2}{M}-\frac{L_1^2}{cM^{2}} &\le \mathbb{E}\big[L_{S_u}(\hat{\phi}_{Q_u})-L_{S_u}(\phi_u^*)\big]\nn \\
   & \le (1+\epsilon_M)\frac{\tau^2}{M}+\frac{R_{\max}}{M^2},
\end{align}
where $\tau^2 \triangleq \frac{1}{2} Tr((I_{Q_u}(\phi_u^*)+\lambda I)^{-1} I_{S_u}(\phi_u^*) )$.
\end{theorem}

\begin{remark}
The proof of Theorem \ref{thm:main} only requires the condition that $\phi_u^*$ is the minimizer of the cross entropy loss $L_{S_u}(\phi_u)$. Thus, similar bounds on different loss function, for example, mean square error, can be obtained using similar proof technique with different $\tau^2$ expressions.
\end{remark}

The upper and lower bounds in Theorem \ref{thm:main} demonstrate that the cross entropy loss of the ML estimator using question set $Q_u$ with size $M$ is essentially $ Tr((I_{Q_u}(\phi_u^*)+\lambda I)^{-1} I_{S_u}(\phi_u^*) )/M$. Motivated by this result, we should select the question set $Q_u$  that minimizes $ Tr((I_{Q_u}(\phi_u^*)+\lambda I)^{-1} I_{S_u}(\phi_u^*) )$. Unfortunately, we cannot do this, since $\phi_u^*$ is unknown. Recall that we use $\hat\phi_{e\mathrm{ML}}$ in the original ML estimator \eqref{eq:MLE_phi} as the true value of $\phi_e$
to calculate $H(\phi_e,\phi_u)$, again we can use $\hat\phi_{u\mathrm{ML}}$ as a coarse approximation of $\phi_u$. We then choose the set $Q_u$ which minimizes $Tr((I_{Q_u}(\hat{\phi}_{u\mathrm{ML}})+\lambda I)^{-1}I_{S_u}(\hat{\phi}_{u\mathrm{ML}}))$. The algorithm is formally presented in Algorithm \ref{alg:main}.
\vspace{-0.1cm}
\begin{algorithm}
   \caption{Fisher information based algorithm}\label{alg:main}
    \begin{algorithmic}
        \State \textbf{Input:} Unlabeled set $S_u$ and labeled set $T_u$, ML estimates of latent vectors $\hat{\phi}_{u\mathrm{ML}}$ and $\hat{\phi}_{e\mathrm{ML}}$ , for $e \in S^{(E)}$
        \State \textbf{Output:} An estimation of latent vector $\hat\phi_{u\mathrm{AL}}$
        \State \textbf{Method:}
        \State 1: Solve the following semi-definite programming (SDP) problem (refer to Theorem \ref{thm:main}):
        \begin{equation*}
          Q_{u}^* = \underset{{Q_u:|Q_u|=M}}\argmin Tr((I_{Q_u}(\hat{\phi}_{u\mathrm{ML}})+\lambda I)^{-1}I_{S_u}(\hat{\phi}_{u\mathrm{ML}}))
        \end{equation*}
        \State 2: Solve the ML estimation on labeled set $Q_u^* \cup T_u$:
            \begin{equation*}
              \hat{\phi}_{u\mathrm{AL}}= \argmin_{\phi_u \in \Lambda} \hat{L}_{Q_u^*\cup T_u}( \phi_{u} )+ \lambda \|\phi_u\|_2^2.
            \end{equation*}
    \end{algorithmic}
\end{algorithm}
\vspace{-0.2cm}
\begin{remark}
To avoid the computational issue of solving SDP when the unlabeled set $S_u$ is extremely large, since $Tr(A^{-1}B)\le Tr(A^{-1})Tr(B)$ holds for positive definite matrices, we can approximate the SDP problem by minimizing $Tr((I_{Q_u}(\hat{\phi}_{u\mathrm{ML}})+\lambda I)^{-1})$ or even maximizing $Tr(I_{Q_u}(\hat{\phi}_{u\mathrm{ML}})+\lambda I)$ as discussed in \cite{SouratiALED17}.
\end{remark}


\section{Experimental setup}
\label{sec:setup}
To demonstrate the efficiency of Algorithm \ref{alg:main} and verify the theoretical results, we compare our Fisher information based active learning algorithm with four baseline algorithms on a real Yelp dataset and an illustrative synthetic dataset. In this section, we introduce these datasets and discuss the setup of our experiments.
\subsection{Yelp Dataset}
\label{sec:yelp}
Yelp contains rich relational data for businesses and users in the form of business categories, user reviews and ratings \cite{gupta2015collective}. Much of this relational data characterizes different levels of user preferences that pose exciting potential for integration, for example, incorporating additional information about the business categories and user preferences can significantly improve rating prediction.
The entities present in the Yelp database are users, businesses and categories. We denote the set of these entities by $S^{(U)}$, $S^{(B)}$ and $S^{(C)}$, respectively and represent each entity by a $k$-dimensional latent vector. In the following subsection, we describe in detail the various relations we use from Yelp and show how we represent them as binary relational matrices.

\subsubsection{Ratings}
The original 5-scale ratings are converted to a binary-valued relation between businesses and users with high ratings (4 and 5) as positive($1$) and low ratings ($\leq 3$) as negative($-1$). We denote the resulting binary rating matrix by $R$ with size $|S^{(B)}|\times |S^{(U)}|$.



\subsubsection{Business Categories}
Each business in the Yelp dataset is categorized according to the nature of the business. The categories available include broad-level classes such as Restaurant, Bank and Grocery, and fine-grained descriptions such Mexican Food, Pizza, and Delis. The business category data can be viewed as a binary relation between businesses and categories and is represented as matrix $BC$ (fully-observed
and complete) with size $|S^{(B)}|\times |S^{(C)}|$.

\vspace{-0.1cm}
\subsubsection{User Categories}
The relation between users and categories are not contained in the Yelp dataset explicitly. Thus, we construct the matrix of this relation using the following way to memorize the user's opinions with respect to a certain type of food: if the user $u$ has visited a business with category $c$, then we set $UC_{uc}=1$, which means the user $u$ holds a positive opinion towards the category $c$.
\begin{figure}[t]
\begin{center}
\includegraphics[width=0.48\textwidth]{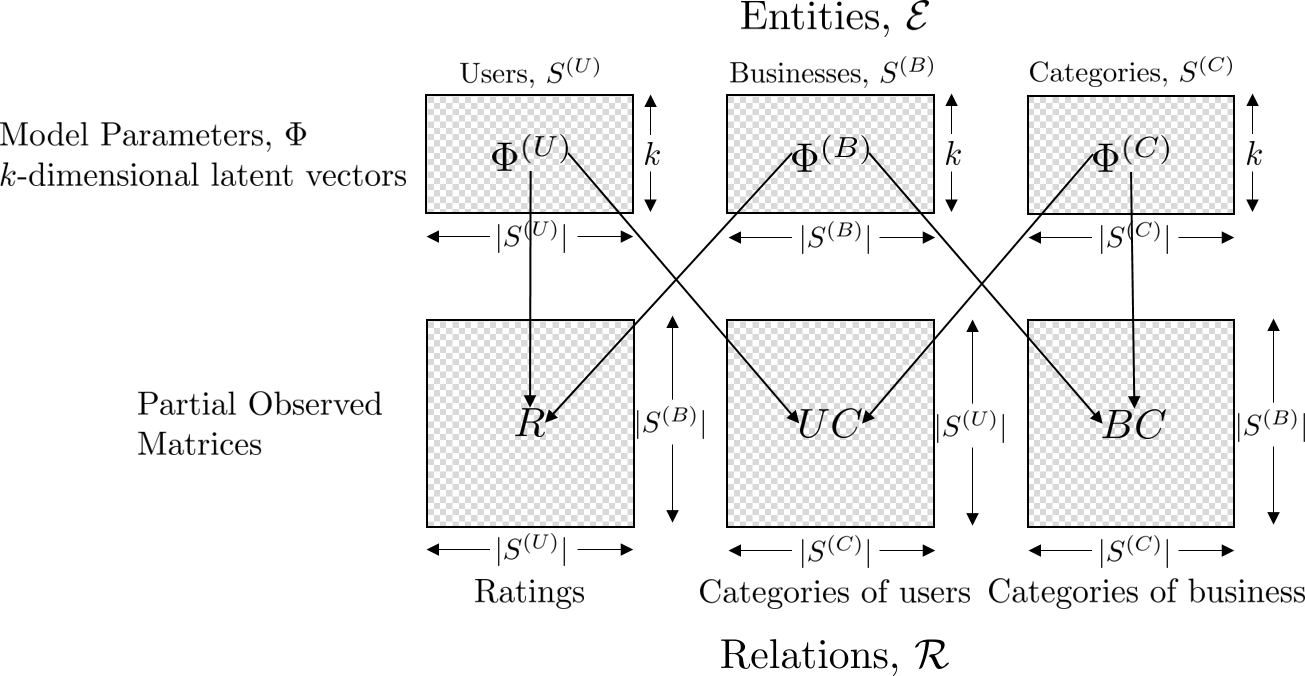}\\
\caption{Overview of the entities and the relations in the collective factorization model for the Yelp Dataset.}
\vspace{-0.6cm}
\label{fig:relation}
\end{center}
\end{figure}
\subsubsection{Datasets and Sizes}
The Yelp dataset contains data from 9 cities across 4 countries; we focus on the data from Urbana-Champaign. For our active learning algorithm comparison, we only consider users that have at least 10 ratings and categories associated with at least 5 businesses.

Since the $UC$ matrix only contains observed categories (all positives), we sample
negative data entries for each user by randomly selecting a set of categories that were not observed. The number of negative samples chosen for each user is same as the number of positive samples.

After this procedure, our modified Yelp dataset contains 473 users, 858 businesses and 33 categories. Note that the ratings matrix $R$ and the user categories matrix $UC$ are very sparse, only $2.6\%$ of $R$ and $24.7\%$ of $UC$ are observed. 

\subsection{CMF Generated Synthetic Dataset}
\label{sec:synthetic_data}
Since the Yelp dataset is not actually generated from the CMF model, it is possible that Theorem \ref{thm:main} does not hold at all and Algorithm \ref{alg:main} cannot achieve a reasonable convergence rate. Thus, we also construct an illustrative synthetic dataset using the CMF model defined in \eqref{eq:CMF} to show the efficiency of our active learning algorithm.

We draw samples of matrices $R$, $BC$, $UC$ using the same structure as we discussed for the Yelp data (Figure \ref{fig:relation}), where $\phi^{(U)}_u, \phi^{(B)}_b, \phi^{(C)}_c$ are generated i.i.d. from the Gaussian distribution $\mathcal{N}(0.25,0.1)$. We set $|S^{(B)}|=|S^{(U)}|=100$, and $|S^{(C)}|=40$, and the dimension of latent vector $k=10$.

\begin{figure*}[!htp]
  \centering
  \subfigure[Yelp data:  matrix $R$ only]{
  \begin{minipage}[b]{0.4\textwidth}
    \centering
    \includegraphics[width=0.9\textwidth]{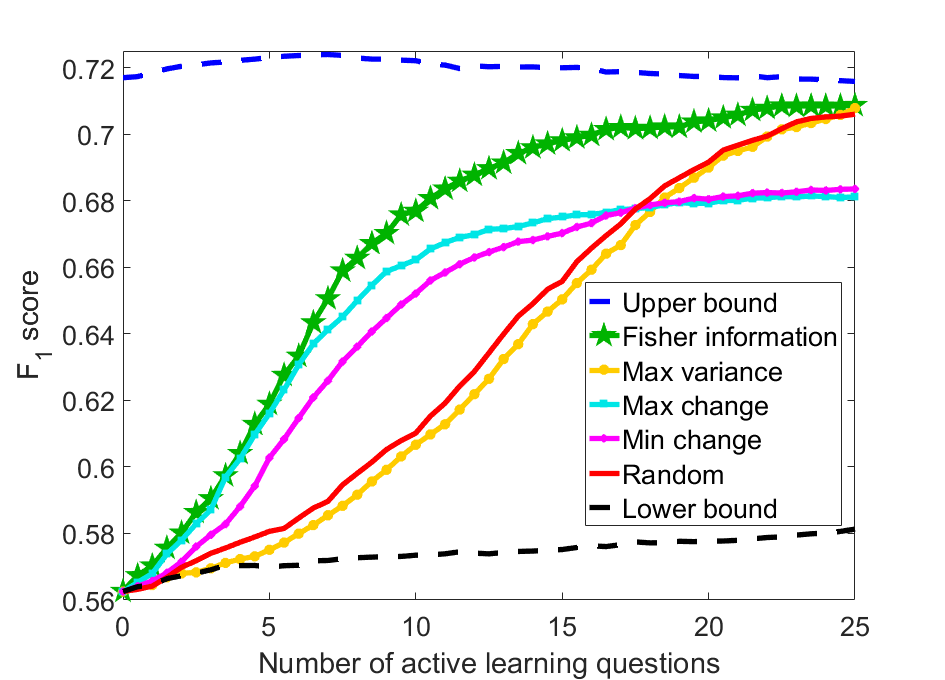}\\
    \label{fig:Yelp_R}
  \end{minipage}}%
  \hspace{0.1\linewidth}%
  \subfigure[Yelp data:  matrices $R$, $BC$ and $UC$]{
  \begin{minipage}[b]{0.4\textwidth}
    \centering
    \includegraphics[width=0.9\textwidth]{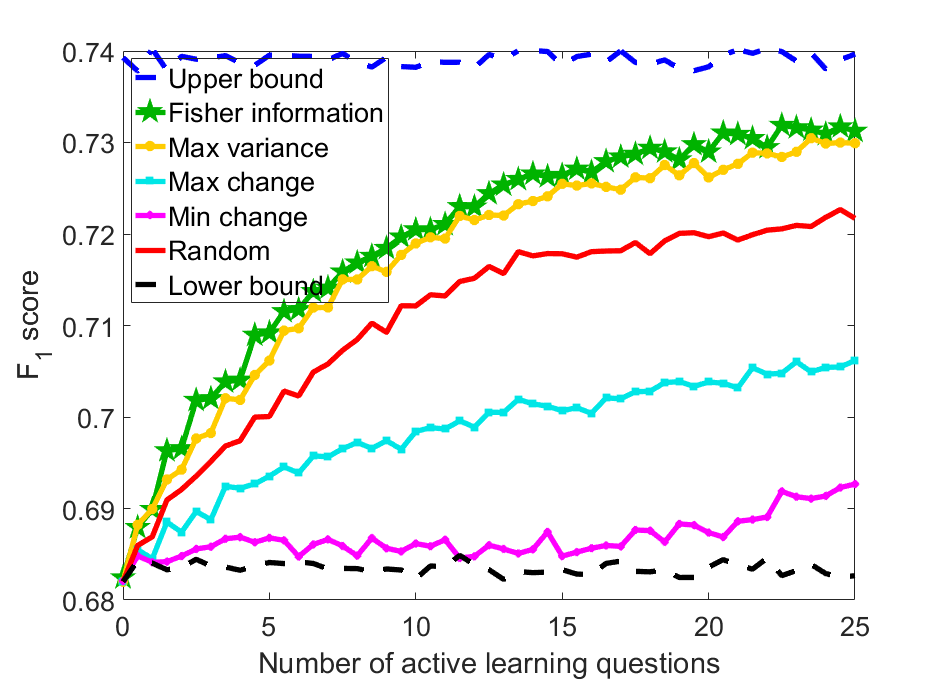}\\
    \label{fig:Yelp_UBC}
  \end{minipage}}\\
  \vspace{-0.3cm}
  \subfigure[Synthetic data:  matrix $R$ only]{
  \begin{minipage}[b]{0.4\textwidth}
    \centering
    \includegraphics[width=0.9\textwidth]{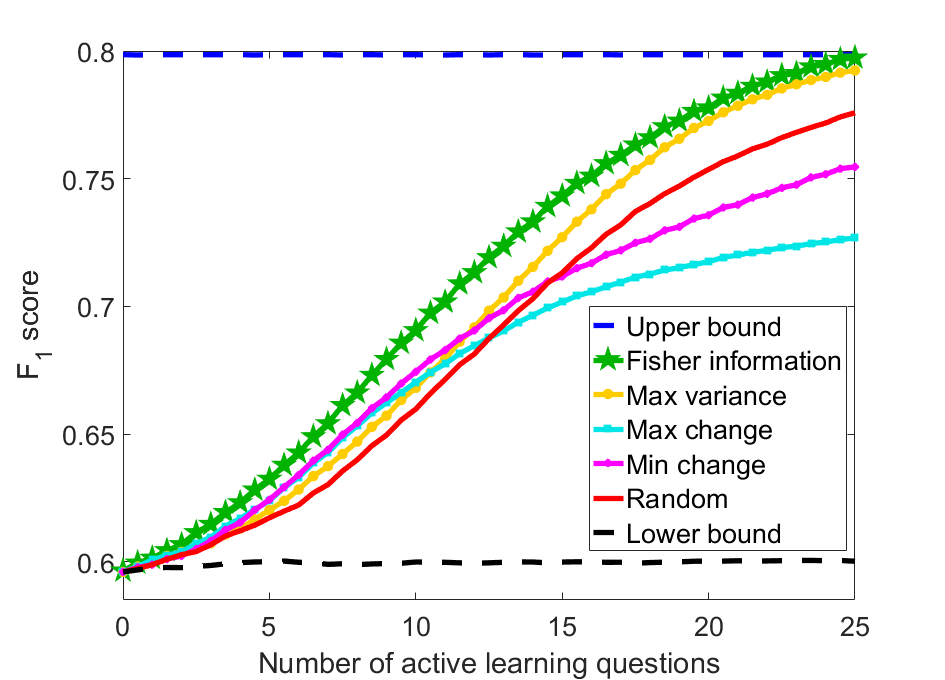}\\
    \label{fig:Simulation_R}
  \end{minipage}}%
  \hspace{0.1\textwidth}
  \subfigure[Synthetic data:  matrices $R$, $BC$ and $UC$]{
  \begin{minipage}[b]{0.4\textwidth}
    \centering
    \includegraphics[width=0.9\textwidth]{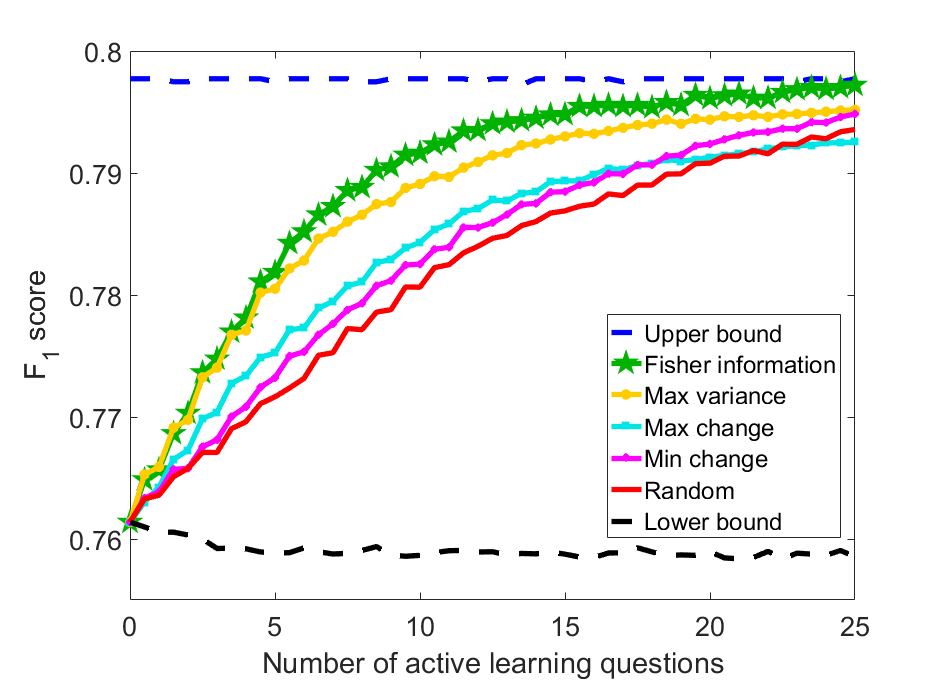}\\
    \label{fig:Simulation_UBC}
  \end{minipage}}
  \vspace{-0.3cm}
  \caption{$F_1$ score comparison of different active learning strategies. Fisher information based algorithm outperforms other algorithms presented in Section \ref{sec:baseline} in all settings. }
\end{figure*}
\subsection{Baseline Algorithms}
\label{sec:baseline}

\textbf{Uncertainty sampling:}
Select the set of entities with the highest noisy affinity variance.

\noindent\textbf{Maximum or Minimum Model Change:} Select the entities that lead to the greatest (smallest) model change $\|\hat\Phi_{\mathrm{ML}}(\mathcal{D})-\hat\Phi_{\mathrm{ML}}(\mathcal{D}\cup Q_u)  \|_F$, if we knew the labels \cite{elahi2016survey}.

%

\noindent\textbf{Random selection:} The $M$ entities are selected randomly from the pool of available question set $S_u$. While ostensibly rather weak, $Q_u$ is actually a statistically good representative
of $S_u$ if sufficiently large number of samples are collected. 

\noindent\textbf{Upper and Lower bound:} These two baselines are not active learning schemes, but they are provided in the following comparison to show the efficiency of active learning. The lower bound is obtained by training the CMF model without any active questions, which is the starting point of all the other algorithms; while the upper bound is obtained by using all the available training samples.

\subsection{Active Learning Setup}
The primary evaluation for comparing different active learning algorithms will be with respect to predicting user ratings $R$. In particular, we consider whether incorporating different level of questions into the active learning process provides significant improvement
in predictions. The simple model that performs the
standard matrix factorization of $R$ will only ask active learning question with respect to matrix $R$.
Moreover, we can compare it with the case where the
model predicts ratings by incorporating business categories, user categories and perform active learning on both $R$ and $UC$.

Since we are unable to retrieve the actual answer from a real user, we need to construct the ground truth for active learning responses. In the following experiments, we use the predictions given by the pre-trained CMF model using all available training data as the ground truth, which also serves as the upper bound of active learning performance. 
We run cross validation to determine the hyper parameter for the pre-trained model and other test models. The value of the regularization constant $\lambda=0.1$, dimension of the latent variables $k=30$ ($k=10$ for Synthetic data) and learning rate $\eta=0.02$ is used. We use the default logistic threshold of $0.5$ for the output of the sigmoid function to predict whether a relation holds between entities. The performance of the prediction is measured in terms of the $F_1$ score defined as the harmonic mean of the
precision and recall. 

\begin{figure*}[t]
  \centering
  \subfigure[Yelp data:  matrix $R$ only]{
  \begin{minipage}[b]{0.4\textwidth}
    \centering
    \includegraphics[width=0.9\textwidth]{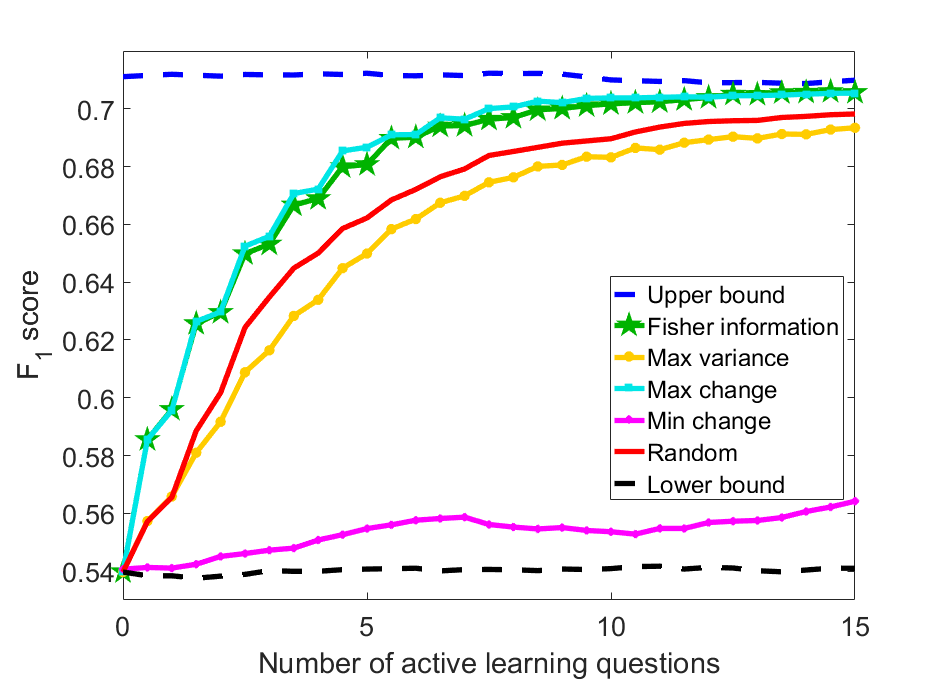}\\
    \label{fig:Yelp_cold_start_R}
  \end{minipage}}%
  \hspace{0.1\linewidth}%
  \subfigure[Synthetic data:  matrix $R$ only]{
  \begin{minipage}[b]{0.4\textwidth}
    \centering
    \includegraphics[width=0.9\textwidth]{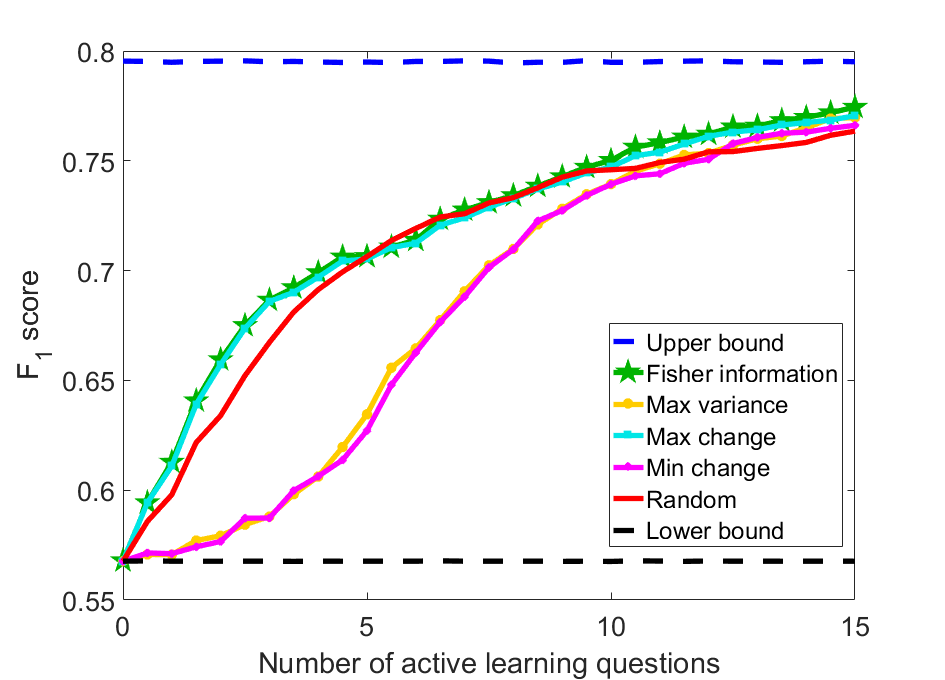}\\
    \label{fig:Simulation_cold_start_R}
  \end{minipage}}
  \vspace{-0.3cm}
  \caption{$F_1$ score comparison of different active learning strategies in the cold start setting. }
\end{figure*}

\section{Experimental Results}
\label{sec:exp}
In this section, we compare the performance of different active learning algorithms with the CMF model in predicting user ratings under different settings. First, we present the rating prediction experiments under the personalized active learning setting in Section \ref{sec:exp_active} on both Yelp and synthetic dataset. Secondly, we investigate the performance of different
active learning algorithms for user cold-start estimation on these two datasets in Section \ref{sec:exp_cold_start}, where we make rating predictions for users with no past observed ratings or reviews.
Finally, for the synthetic dataset, we compare the performance of different active learning strategies when the model is updated with noisy responses in Section \ref{sec:noise_exp}.



\subsection{Personalized Active Learning}
\label{sec:exp_active}

To compare the performance of different active learning strategies in the personalized active learning setting, we perform evaluations on a held-out
test set from the observed data. The process can be described as follow:
\begin{enumerate}
  \item For each user in the dataset, we randomly choose 20\% (30\% for synthetic data, the same below) from the original matrix $R$ and $UC$ as the test data, 20\% (10\%) as the training set, and perform active learning on the remaining 60\% of data.
  \item In each iteration, we randomly choose 25\% of users, and ask them one question based on different question selection algorithm. The CMF model is then updated with answers estimated by the pre-trained model using 80\% (70\%) of all available training data.
  \item Repeat this active learning process for 25 iterations.
  \item {Compute the $F_1$ score on the test dataset with 50 trials of Monte Carlo runs}.
\end{enumerate}

In Figures \ref{fig:Yelp_R} and \ref{fig:Yelp_UBC}, we compare Algorithm \ref{alg:main} with the four active learning baselines described in Section \ref{sec:baseline} on the Urbana-Champaign dataset. Figure \ref{fig:Yelp_R} shows the performances of models trained using exclusively the ratings matrix $R$ and the models in Figure \ref{fig:Yelp_UBC} are trained using matrices $R$, $BC$ and $UC$ collectively. Moreover, Figure \ref{fig:Simulation_R} and \ref{fig:Simulation_UBC} show experimental results on the synthetic dataset.

We first observe that $F_1$ scores achieved by the lower bound using only $R$ in Figures \ref{fig:Yelp_R} and \ref{fig:Simulation_R} are $0.58$ and $0.60$ respectively. The lower bound in Figures \ref{fig:Yelp_UBC} and \ref{fig:Simulation_UBC} using $R$, $BC$ and $UC$ are $0.685$ and $0.760$ respectively, showing that incorporating different types of information significantly improves the prediction accuracy.

We also note that the upper bound performance can be achieved with fewer samples by using active learning. Specifically, in Figures \ref{fig:Yelp_R} and \ref{fig:Yelp_UBC}, the upper bound is trained with $\sim8500$ ratings (80\% of all data), but the active learning algorithms requires $\sim5000$ ratings (start from 20\% of all data, combined with 25 rounds of updates). The data efficiency improvements can also be observed in  Figures \ref{fig:Simulation_R} and \ref{fig:Simulation_UBC} for the synthetic dataset.

Moreover, our Fisher information based algorithm is strictly better than all the other baseline algorithms in all cases. This improvement of Fisher information based algorithm is smaller when using $R$, $BC$ and $UC$ collectively, since by utilizing different levels of information, the lower bound achieved in this setting is quite promising, which limits the potential improvements of active learning.


\subsection{Cold Start Setting}
\label{sec:exp_cold_start}
The other major challenges faced by recommendation
systems is to predict ratings for new users for which no reviews or ratings have been observed. To compare the performance of different active learning strategies in the cold-start setting, we carry out the following experiment:
\begin{enumerate}
  \item We randomly choose 20\% (40\%) of users as the cold start users, the training dataset contains no records for these cold-star users. The test set is constructed with 50\% cold start users data.
  \item In each iteration, we ask one active learning question to all cold-start users based different question selection algorithm. The CMF model is then updated with answers estimated by the pre-trained model using 50\% of available training data for cold-start users.
  \item Repeat this active learning process for 15 rounds.
  \item Compute the $F_1$ score on the test dataset for the cold-start users with 50 trials of Monte Carlo runs.
\end{enumerate}

In the Figure \ref{fig:Yelp_cold_start_R} and \ref{fig:Simulation_cold_start_R}, we compare our Fisher information based active learning algorithm in the cold start setting using both the Yelp dataset and the synthetic dataset.

Since there is no records about the cold-start users in the training set, the lower bound in these figures start from the $F_1$ score of $0.5$ with purely guessing. However, a reasonable $F_1$ score of $0.65$ can be achieved after just 2 queries with Algorithm \ref{alg:main} in both figures, which demonstrates the applicability of our results even in the cold-start setting.

We note that the Maximum Model Change baseline works quite well in Figure \ref{fig:Yelp_cold_start_R} and \ref{fig:Simulation_cold_start_R}, one possible interpretation is that exploring user preference is more important than minimizing the noise of response in the cold-start setting. However, note that our proposed method doesn't require changing querying functions between the cold-start and general setting as it achieves strong performance in both cases.

\vspace{-0.1cm}
\subsection{Noise Tolerance}
\label{sec:noise_exp}
\begin{figure}[!h]
    \centering
    \includegraphics[width=0.36\textwidth]{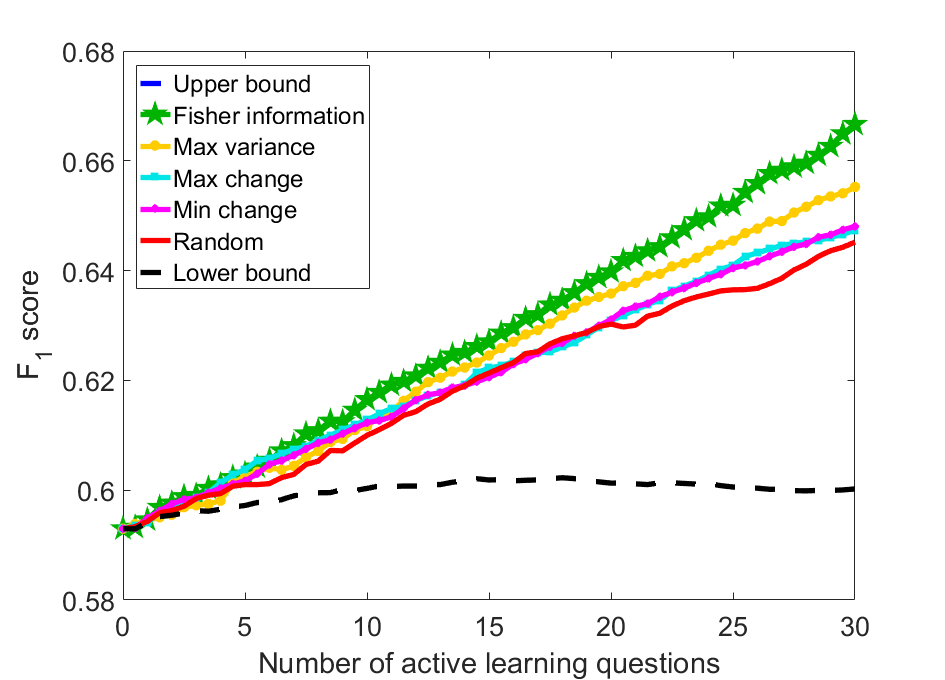}\\
    \caption{$F_1$ score comparison of different active learning strategies with noisy answers. Upper bound is around 0.80.}
    \label{fig:Noisy_R}
\end{figure}

To study the performance of different active learning strategies when only noisy responses are available, we modify the experiment described in Section \ref{sec:exp_active}. Instead of updating the model with answers estimated by the pre-trained CMF model, we regenerate samples using the same distribution used to construct the synthetic dataset in Section \ref{sec:synthetic_data}. Thus, the user response to the same question from different Monte Carlo runs will be flipped, with results shown in Figure \ref{fig:Noisy_R}.

Comparing Figure \ref{fig:Noisy_R} with Figure \ref{fig:Simulation_R}, the performance of all the algorithms decrease due to the influence of the noisy response. The proposed Fisher information based algorithm still outperforms all the others in this situation, demonstrating the robustness of our method.

\section{Conclusions}
\label{sec:conclusion}
In this paper, we consider the question selection problem for recommendation systems with multi-level user preferences. Building on the CMF framework, we provide a theoretical analysis of an optimal active learning strategy with a realizable approximation algorithm. Our experiments on synthetic and Yelp data demonstrate that the proposed algorithm performs well in different practical settings. In the future, we plan to address the problem of noisy response in active learning, especially when the quality of response is related to the type of questions.  This would be particularly relevant in the context of dialogue managers for modeling the {\em natural language generation} $\rightarrow$ {\em user} $\rightarrow$ {\em natural language understanding} loop within conversational recommendation systems in a more abstract setting.


\clearpage
\small
\bibliographystyle{aaai}
\bibliography{IConA-KDD}

\clearpage

\appendix

\section{Regularity conditions for Asymptotic Normality of ML estimator}
\label{appx:assum}
The following regularity conditions are needed to establish the standard Asymptotic Normality of ML estimation \cite{ferguson2017course}.

\begin{assump}\label{assump:LAN}
(Regularity conditions for ML estimation)
\begin{enumerate}
  \item \textbf{Identifiability}: $\ell(y|\phi_e,\phi_u) \ne \ell(y|\phi_e',\phi_u') $, for $\Phi \ne \Phi'$.
  \item \textbf{Compactness}:  $\Phi^*$ is an interior point of the compact set $\Lambda^{k \times |\mathcal{E}|}$.
  \item \textbf{Smoothness}: $\ell(y|\phi_e,\phi_u)$ is smooth, such that the first, second and third
derivatives of $\phi_e,\ \phi_u$ exist.
  \item \textbf{Strong Convexity}: The Fisher information matrix of $\ell(y|\phi_e,\phi_u)$ is positive definite.
  \item \textbf{Boundedness}: $\max_{y} \sup_{\Phi}|\ell(y|\phi_e,\phi_u)|\le R_{\max}$.
\end{enumerate}
\end{assump}



Note that in the original CMF model \eqref{eq:CMF}, parameter $\Phi$ is not unique and the parameter space is not compact. We avoid this problem by focusing on the $\ell_2$ regularized estimation in \eqref{eq:MLE_phi}, and further assume that $\Phi^* = \argmin_{\Phi\in \Lambda^{|\mathcal{E}|}} \big( L_{\mathcal{D}}(\Phi) + \lambda \|\Phi\|_F^2 \big)$. Thus, it can be verified that Assumption \ref{assump:LAN} holds for \eqref{eq:MLE_phi}, which ensures the asymptotic normality.

\section{Useful lemma}
\label{appx:lemmas}
\begin{lemm} \cite{ChaudhuriKNS15} \label{lemma:main}
Suppose $\psi_1(\theta),\cdots,\psi_n(\theta):\mathbb{R}^d\to \mathbb{R}$ are random functions drawn i.i.d. from a distribution, where $\theta \in \mathcal{S} \subseteq \mathbb{R}^d$. Denote $P(\theta) = \mathbb{E}[\psi(\theta)]$ and $Q(\theta):\mathbb{R}^d\to \mathbb{R}$ be another function. Let  $\hat{\theta} = \argmin_{\theta\in \mathcal{S}} \sum_{i=1}^n \psi_i(\theta)$, and $\theta^*=\argmin_{\theta\in \mathcal{S}}P(\theta)$.
Assume:
\begin{enumerate}
  \item Assumption \ref{assump:LAN} holds for $\psi(\theta)$.
  \item Assumption \ref{assump:opt} holds for $P(\theta)$ and $\psi(\theta)$.
  \item For $Q(\theta)$, we need  $\nabla Q(\theta^*)=0$.
  \item There exists a neighborhood $B$ of $\theta^*$ and a constant $L_3$, such that $\nabla^2 Q(\theta)$ are $L_3$-Lipschitz, namely
    \begin{align*}
      \Big\|\big(\nabla^2Q(\theta^*)\big)^{-1/2}\big(\nabla^2 Q(\theta)- &\nabla^2 Q(\theta')\big)\big(\nabla^2Q(\theta^*)\big)^{-1/2}\Big\| \\
                    &\le L_3\|\theta-\theta'\|_{\nabla^2P(\theta^*)},
    \end{align*}
    hold with probability one, for $\theta,\theta' \in B$.
\end{enumerate}
Choose $p \ge 2$ and define $\epsilon_n \triangleq  c(L_1 L_3+\sqrt{L_2}) \sqrt{\frac{p\log dn}{n}}$, where $c$ is an appropriately chosen constant. Let $c'$ be another appropriately chosen constant. If $n$
is large enough so that $\sqrt{\frac{p\log dn}{n}}\le c' \min\left\{\frac{1}{\sqrt{L_2}},\frac{1}{L_1L_3},\frac{\mathrm{diameter}(B)}{L_1} \right\}$, then:
\begin{align*}
  (1-\epsilon_n)\frac{\tau^2}{n}-\frac{L_1^2}{n^{p/2}} &\le \mathbb{E}\big[Q(\hat{\theta})-Q(\theta^*)\big] \\
   & \le (1+\epsilon_n)\frac{\tau^2}{n}+\frac{\max_{\theta\in S}Q(\theta)-Q(\theta^*)}{n^p},
\end{align*}
where $\tau^2 \triangleq \frac{1}{2} Tr\Big(\nabla^2Q(\theta^*) \big(\nabla^2P(\theta^*)\big)^{-1}\Big)$.
\end{lemm}

\section{Proof Sketch for Theorem \ref{thm:main}}
\label{appx:thm1}
We first let $\theta=\phi_u$, $\theta^* = \phi_u^*$, then
\begin{equation}
  \psi_i(\theta) = \ell(Y|\phi_i, \phi_u )+ \lambda\|\phi_u\|_2^2,
\end{equation}
where $\phi_i \in Q_u$ and $Y \sim P_{\Phi^*}(Y|\phi_i, \phi_u^* )$, for $i=1,\cdots, |Q_u|$, then $P(\theta) = L_{Q_u}(\phi_u)+ \lambda \|\phi_u\|_2^2$ and let $Q(\theta) = L_{S_u}(\phi_u)$.
Using the notation in Section \ref{subsec:bound}, we can compute that
\begin{align*}
  \nabla^2P(\theta^*) &= \nabla^2 \Big( L_{Q_u}(\phi_u^*)+ \lambda \|\phi_u^*\|_2^2 \Big) = I_{Q_u}(\phi_u^*)+\lambda I,\\
  \nabla^2Q(\theta^*) &= \nabla^2 \Big(L_{S_u}(\phi_u^*) \Big) = I_{S_u}(\phi_u^*).
\end{align*}
Using the Assumption \ref{assump:opt} and the condition in Theorem \ref{thm:main} that $I_{Q_u}(\phi_u^*) \succeq c I_{S_u}(\phi_u^*)$, we see
that this satisfies the hypothesis of Lemma \ref{lemma:main} with constants
\begin{equation*}
  (\tilde{L}_1,\tilde{L}_2,\tilde{L}_3) = (L_1/\sqrt{c}, L_2/c,L_3/c^{3/2}).
\end{equation*}
We now apply Lemma \ref{lemma:main} to prove Theorem \ref{thm:main} and it can be verified that
\begin{align}
\tau^2 
  = \frac{1}{2}Tr((I_{Q_u}(\phi_u^*)+\lambda I)^{-1} I_{S_u}(\phi_u^*) ).
\end{align}

\end{document}